\documentclass[conference]{IEEEtran}
\IEEEoverridecommandlockouts
\usepackage{cite}
\usepackage{amsmath}
\usepackage{amsfonts,amssymb,amsfonts}
\usepackage[pdftex]{graphicx}
\usepackage{algorithmic}
\usepackage{textcomp}
\def\BibTeX{{\rm B\kern-.05em{\sc i\kern-.025em b}\kern-.08em
    T\kern-.1667em\lower.7ex\hbox{E}\kern-.125emX}}

\newcommand{\bU}{\mathbf U}
\newcommand{\bW}{\mathbf W}
\newcommand{\bb}{\mathbf b}
\newcommand{\bx}{\mathbf x}
\newcommand{\bh}{\mathbf h}

\newtheorem{defn}{Definition}[section]
\newtheorem{rem}[defn]{Remark}

\title{Recurrent Neural Networks for\\
       Learning Long-term Temporal Dependencies with
       Reanalysis of Time Scale Representation}

\author{\IEEEauthorblockN{Kentaro Ohno, Atsutoshi Kumagai}
\IEEEauthorblockA{
NTT Computer and Data Science Laboratories \\
\{kentaro.ohno.tf, atsutoshi.kumagai.ht\}@hco.ntt.co.jp
}
}

\begin{document}
    \maketitle 

    \begin{abstract}
        Recurrent neural networks with a gating mechanism such as
        an LSTM or GRU are powerful tools to model sequential data.
        In the mechanism, a forget gate,
        which was introduced to control information flow in a hidden state in the RNN,
        has recently been re-interpreted as a representative of 
        the time scale of the state,
        i.e., a measure how long the RNN retains information on inputs.
        On the basis of this interpretation, several parameter initialization methods 
        to exploit prior knowledge on temporal dependencies in data 
        have been proposed to improve learnability.
        However, the interpretation relies on various unrealistic assumptions,
        such as that there are no inputs after a certain time point.
        In this work, we reconsider this interpretation of the forget gate in a more realistic setting. 
        We first generalize the existing theory on gated RNNs 
        so that we can consider the case where inputs are successively given. 
        We then argue that the interpretation of a forget gate as a temporal representation is valid 
        when the gradient of loss with respect to the state decreases exponentially as time goes back. 
        We empirically demonstrate that existing RNNs satisfy this gradient condition 
        at the initial training phase on several tasks, 
        which is in good agreement with previous initialization methods. 
        On the basis of this finding, we propose an approach to construct new RNNs 
        that can represent a longer time scale than conventional models, 
        which will improve the learnability for long-term sequential data. 
        We verify the effectiveness of our method by experiments with real-world datasets.
    \end{abstract}

    \section{Introduction}
    Recurrent Neural Networks (RNNs) are deep learning models for representing sequential data,
    which have extensive applications including speech recognition \cite{amodei2016deep,graves2013speech},
    natural language processing \cite{sutskever2014sequence}, video analysis \cite{yue2015beyond}, 
    and action recognition \cite{ordonez2016deep}.
    RNNs represent the temporal features of data using time-variant hidden states
    whose transition is determined by the previous state and an input at the present time.
    RNNs are typically trained by gradient descent methods using backpropagation through time.
    However, the training is usually difficult because the gradient of loss
    tends to take too small value as the sequence length increases.
    which is known as the vanishing gradient problem \cite{bengio1994learning,pascanu2013difficulty}.
    To enable models to learn on long-term sequential data,
    RNNs with a gating mechanism (called gated RNNs), such as 
    a Long Short-Term Memory (LSTM) \cite{hochreiter1997long} or Gated Recurrent Unit (GRU) \cite{cho2014learning}, 
    have been proposed.
    Gated RNNs control how much information of the past state 
    is retained to the next state by means of a forget gate function \cite{gers1999learning},
    which is useful to mitigate the vanishing gradient problem 
    \cite{van2018unreasonable}.
    Furthermore, the forget gate has recently been considered to take
    a role to represent a temporal characteristic in RNN models \cite{tallec2018can}.
    That is, output values of the forget gate can be viewed as an expression
    of how long the state keeps information (or {\it memory}),
    which is called the {\it time scale} of the state\footnote{Note
        that the idea is considered as an extension of temporal representation in older RNN models
        such as leaky units \cite{mozer1992induction,jaeger2002tutorial}.
        }.
    On the basis of this interpretation, several methods to impose a desired time scale 
    on gated RNNs have been proposed \cite{tallec2018can,gua2020improving,mahto2020multi}.

\begin{figure*}[t]
        \begin{minipage}{0.33\hsize} 
            \centering
            \includegraphics[width=0.29\paperwidth]{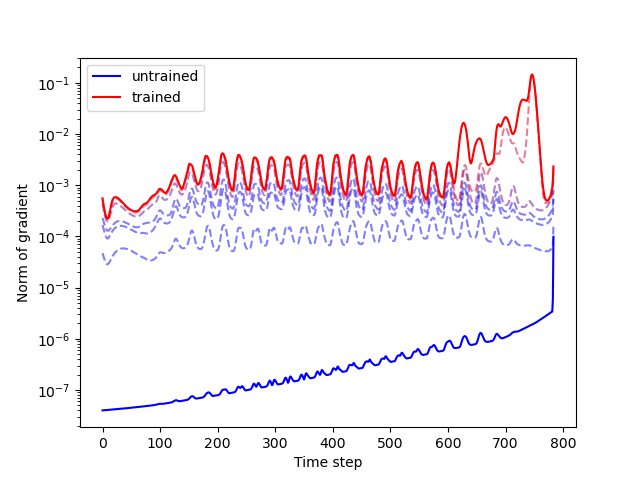}
        \end{minipage}
        \begin{minipage}{0.33\hsize}            
            \centering
            \includegraphics[width=0.29\paperwidth]{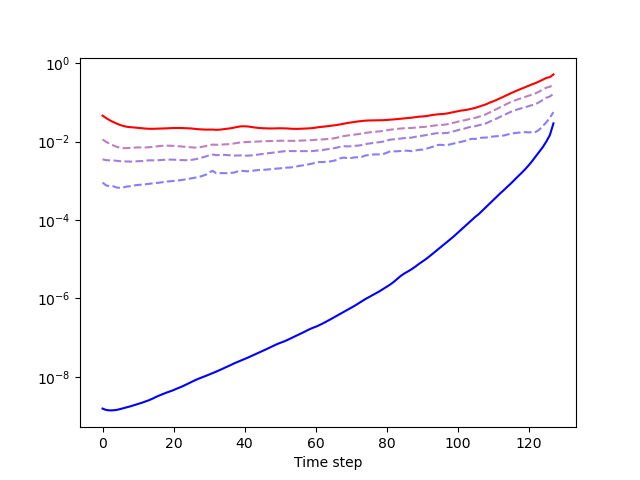}
        \end{minipage}  
        \begin{minipage}{0.33\hsize}            
            \centering
            \includegraphics[width=0.29\paperwidth]{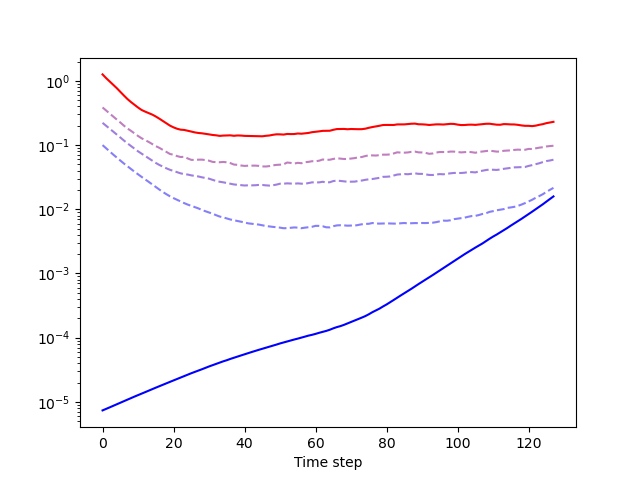}
        \end{minipage}  
        \caption{
            Euclidean norm of gradient of the cross-entropy loss with respect to 
            an input at each time step $||\partial L / \partial \mathbf x_t||$.
            An LSTM on sMNIST (left) and HAR (center) tasks (Section \ref{sec:experiment})
            and a leaky RNN on HAR (right) are shown.
            Gradient is backpropagated from right to left over time.
            At initialization (blue solid line), the gradient decrease exponentially along with back-propagation.
            After learning (red solid line), there is no longer any such exponential behavior.
            Dashed lines correspond to the model after 1, 5, and 10 training epochs,
            from below (blue) to upper (red) 
            with additional plots after 50 and 100 epochs in the left figure.
        }
        \label{fig:grad_change}
    \end{figure*}

    However, justification of the interpretation
    has not been fully explored.
    The interpretation is commonly explained by a theory using a ``free input" regime \cite{tallec2018can},
    which ignores inputs after some time step and even parameters in RNNs.
    We empirically found such simplification sometimes generate a gap between
    theoretical properties on a gated RNN and its actual behavior.
    For example, while existing studies indicate that the gradient of loss with respect to 
    inputs decrease exponentially as time goes back in gated RNNs
    \cite{van2018unreasonable,le2019Learning},
    such experimental behavior does not necessarily occur in a trained model (Figure \ref{fig:grad_change}).
    It is important to clarify when and how we can fill this gap
    for a more advanced understanding on RNNs and the construction of more sophisticated models.

    In this paper, we first extend the aforementioned theory
    to make it applicable to more practical situations where inputs are successively given.
    This approach relates the time scale in gated RNNs to the vanishing gradient problem.
    Specifically, we show that when the gradient of loss with respect to inputs decays
    exponentially as time goes back,
    the forget gate indeed represents the time scale of the state.
    Through experimental observation on RNNs trained with several tasks,
    we found that this condition on the gradient holds, at least at the initial phase of training.
    This is a new aspect to explain the effectiveness of recent initialization methods
    \cite{tallec2018can,gua2020improving}
    based on the time scale interpretation of the forget gate function.
    That is, since the theory behind such methods is valid at least at the initial phase of training,
    imposing a desirable time scale on the model at initialization is reasonable.

    On the basis of this observation, we consider a different approach from the existing methods
    to improve the learnability of RNNs at the initial training phase.
    That is, we propose a method to construct new RNN models,
    which can represent a larger order time scale than conventional leaky or gated RNNs.
    Our strategy is to replace the exponential decay of the gradient at
    the initial phase of training with a more gradual (polynomial) decay 
    by changing the model structure.
    By this modification, we expect that the RNNs can capture the long-term dependencies
    of data more efficiently.
    To achieve this, we adopt the differential equation view on RNNs, which is 
    often used in theoretical analysis on RNNs \cite{tallec2018can}.
    The exponential decay of memory in the RNN state is characterized by the solution $h(t) = e^{-c(t-t_0)}h(t_0)$ of 
    a basic ordinary differential equation 
    \begin{align}\label{eq:linear-ode}
        \frac{\mathrm{d}h}{\mathrm{d}t} (t) = -c h(t)
    \end{align}
    with constant $c>0$,
    which is obtained by simplifying the continuous counterpart of state update rule in RNNs.
    By exploiting the fact that modifying the linear decay term in (\ref{eq:linear-ode}) to
    a higher degree term leads to a polynomial decaying solution,
    we can model memory that decays much more slowly than exponentially.
    We add the higher degree term into existing RNNs to derive 
    a new family of models that can represent a much longer time scale.
    This method can be implemented without any additional parameters
    and with little computational overhead.
    We evaluated our method with experiments on 
    sequence classification tasks on real-world datasets
    and found that it indeed helps RNNs to improve the accuracy.

    Our major contributions are as follows.
    \begin{itemize}
        \item We extend the existing arguments \cite{tallec2018can} to 
              establish a theory on temporal structures in gating mechanisms and leaky units,
              which enables us to see
              how time scale in RNNs behaves in more practical situations.
        \item Through observations on models while training, we found that 
              the time scale interpretation in the forget gate is indeed valid, at least at initialization.
              This gives us a new insight on previously proposed initialization methods
              that impose a specific time scale on RNN models at initialization \cite{tallec2018can,gua2020improving}.
        \item On the basis of this insight, we derive a simple method to construct new RNN models
              that can represents a larger order time scale than previous models.
              We experimentally verify our method on real-world datasets.
    \end{itemize}

    In Section \ref{sec:related} of this paper, we briefly review related works.
    Section \ref{sec:preliminaries} explains the existing theory
    on time scale representation in leaky or gated RNNs.
    We extend the theory and compare the theoretical and
    actual behavior of RNNs in Section \ref{sec:extention}.
    We propose our approach to construct RNNs that represent a larger time scale
    in Section \ref{sec:proposal}.
    Section \ref{sec:experiment} verifies our proposed method by experiments.
    We conclude in Section \ref{sec:conclusion} with a brief summary.

    \textbf{Notation.}
    $\mathbb R$ is a set of real numbers.
    Vectors and matrices are denoted by small and capital bold letters (e.g., 
    $\mathbf{x}\in \mathbb{R}^n$ and $\mathbf{W} \in \mathbb{R}^{n\times n}$).
    $\mathbf W^\top$ denotes the transpose of a matrix $\mathbf W$.
    The Jacobian matrix of a vector-valued function $\mathbf y = \mathbf{y}(\mathbf{x})$ 
    at $\mathbf x_0$ is $\frac{\partial \mathbf{y}}{\partial \mathbf{x}}(\mathbf x_0)$,
    or just $\frac{\partial \mathbf{y}}{\partial \mathbf{x}}$ when it does not cause confusion.
    $\mathbf a\odot \mathbf{b} = (a_i\cdot b_i)_i \in \mathbb{R}^n$ denotes the element-wise multiplication, 
    and $\operatorname{diag}(\mathbf{a})\in \mathbb{R}^{n\times n}$ is a diagonal matrix 
    with its diagonal entries equal to the entries of $\mathbf a$.
    A power $\mathbf x^r$ of a vector $\mathbf x$ is taken entry-wise.
    $\mathbf 1 \in \mathbb R^n$ denotes a vector with all entries 1.
    $\mathbf I \in \mathbb R^{n \times n}$ denotes an identity matrix.
    $\sigma(x)=\frac{1}{1+e^{-x}}$ denotes a sigmoid function.
    $f'$ denotes the derivative of a function $f$.

    \section{Related works}\label{sec:related}

    There have been many works on time scale
    representation in RNNs
    \cite{el1996hierarchical,koutnik2014clockwork,liu2015multi,chung2016hierarchical,tallec2018can}.
    Since the use of gated RNNs such as LSTMs and GRUs is
    dominant in a wide range of applications among other RNNs,
    investigating the time scale representation in gated RNNs is
    of particular importance.
    Tallec and Ollivier \cite{tallec2018can} argued that
    the forget gate function in LSTMs and GRUs represents 
    how long memory is retained in hidden states, i.e., the time scale of the states.
    Based on this idea, several methods have been proposed to 
    improve learning ability of gated RNNs
    \cite{tallec2018can,gua2020improving,mahto2020multi}.
    Despite the effectiveness of these methods, 
    a possible gap between the theoretical and actual behavior of hidden states
    in RNNs has been not well understood.
    We aim to take an approach that bridges this gap.
    Moreover, previous applications of the theory have been limited
    to controlling the time scale in states with a bias term in the forget gate
    \cite{tallec2018can,gua2020improving,mahto2020multi}.
    In particular, RNNs used in such methods model memory that decays exponentially. 
    In this work, we establish a method to represent larger order time scales 
    by modeling memory that decays polynomially.
    There is a recent work that incorporates polynomial memory decay into RNNs 
    from a statistical view \cite{zhao2020rnn}.
    However, the method requires to store numerous past states to update the state,
    which increases computational costs for training and inference.
    In contrast, our method does not need to store past states while inference,
    and so has less additional computational costs.
    
    \section{Preliminaries: Time scale in RNNs}\label{sec:preliminaries}

    An RNN consists of a hidden state and its transition driven by inputs.
    Let $\mathbf h_t \in \mathbb R^n$ and $\mathbf x_t \in \mathbb R^d$
    denote a state and an input at time $t$ respectively.
    In order to model complex temporal dependencies in sequential data, 
    it is crucial for RNNs to learn temporal representation efficiently.
    In this section, we review how such representation is implemented in RNN structures
    following Tallec and Ollivier's argument \cite{tallec2018can}.

    We begin with explaining a leaky RNN model \cite{mozer1992induction,jaeger2002tutorial},
    which is a simplified variant of gated RNNs such as LSTMs and GRUs.
    In the leaky RNN, temporal representation is achieved in the simplest way with
    leaky units \cite{mozer1992induction,jaeger2002tutorial}, whose state update rule is written as
    \begin{align}\label{eq:LeakyRNN}
        \mathbf h_t &= (1-\alpha) \mathbf h_{t-1} + \alpha \tilde{\mathbf h}_t, \\
        \tilde{\mathbf h}_t &= \tanh(\mathbf U \mathbf h_{t-1} + \mathbf W \mathbf x_t + \mathbf b),
    \end{align}
    where $\alpha \in (0,1), \mathbf U\in \mathbb R^{n\times n}, \mathbf W \in \mathbb R^{n\times d},$
    and $\mathbf b \in \mathbb R^n$ are parameters to be learned.
    We call $\mathbf U$ a recurrent weight matrix.
    $\alpha$ controls the information flow in the state.
    When $\alpha$ is small, the state is almost unchanged and information is retained,
    and when $\alpha$ is close to 1, the state is largely replaced by a new state $\tilde {\mathbf h}_t$.

    Tallec and Ollivier \cite{tallec2018can} estimated how long the state of a leaky RNN retains information via
    a ``free input" regime, that is, the case where $\mathbf x_t = 0$ for $t>t_0$ for some time step $t_0$,
    with furthur simplification assuming $\mathbf U=0, \mathbf b = 0$. 
    Under this assumption, (\ref{eq:LeakyRNN}) reduces to
    \begin{align}
        \mathbf h_t = (1-\alpha) \mathbf h_{t-1},
    \end{align}
    which leads to an explicit expression 
    \begin{align}\label{eq:simple-decay}
        \mathbf h_t = (1-\alpha)^{t-t_0} \mathbf h_{t_0}.
    \end{align}
    This formula indicates that the state decays by a constant factor
    after every $\tau = - 1 / \log (1-\alpha)$ time steps\footnote{In
    the context of a continuous counterpart of RNN, which was originally argued
    \cite{tallec2018can}, the characteristic time is $\tau = 1/\alpha$. 
    In an asymptotic sense where $\alpha$ is small, these two are equivalent since we have 
    $- \log (1-\alpha) \approx \alpha $.}.
    The decay of the state in (\ref{eq:simple-decay}) is considered as a loss of memory in RNN,
    and the decay rate represents the {\it time scale} of the state, that is, 
    how long the state retains information.

    The above insight can also be applied to RNNs with gating mechanism such as
    an LSTM \cite{hochreiter1997long} or GRU \cite{cho2014learning},
    which is written in a general form as
    \begin{align}\label{eq:gatedRNN}
        \mathbf h_t &= \mathbf f_t \odot \mathbf h_{t-1} + \mathbf i_t \odot \tilde {\mathbf h}_t,   \\
        \mathbf f_t &= \sigma (\mathbf U_f \mathbf h_{t-1} + \mathbf W_f \mathbf x_t + \mathbf b_f), \\
        \mathbf i_t &= \sigma (\mathbf U_i \mathbf h_{t-1} + \mathbf W_i \mathbf x_t + \mathbf b_i).
    \end{align}
    Here, $\tilde {\mathbf h}_t$ is a quantity to represent new information at time $t$.
    $\mathbf f_t$ and $\mathbf i_t$ are called a forget gate and an input gate, respectively.
    Again, by ignoring $\tilde{\mathbf h}_t$ and assuming $\mathbf f_t=\mathbf f$ to be constant,
    we see the same exponential decay structure $\mathbf h_t = \mathbf f^{t-t_0} \odot \mathbf h_{t_0}$.
    Thus, it is considered that the forget gate controls the unit-wise 
    forgetting time to represent complex time scale dynamically \cite{tallec2018can}.
    
    The above theory on time scale in leaky or gated RNNs has been utilized 
    to improve learnability \cite{kusupati2019fastgrnn,tallec2018can,gua2020improving,mahto2020multi}.
    For a leaky RNN, it has been theoretically and empirically shown that 
    setting $\alpha$ proportionally to $1/T$ for sequence length $T$ helps
    with the convergence and generalization of learning \cite{kusupati2019fastgrnn}.
    For gated RNNs,
    when time-variant terms $\mathbf U_f \mathbf h_{t-1}, \mathbf W_f \mathbf x_t$
    in the forget gate $\mathbf f_t$ are centered around zero, 
    $\mathbf f_t$ takes values near $\sigma (\bb_f)$.
    Therefore, the time scale in gated RNNs is considered 
    to be controlled largely by the bias term $\bb_f$.
    In particular, for a gated RNN to learn long-term dependencies efficiently,
    one can impose some entries in $\bb_f$ to take large values 
    at initialization \cite{tallec2018can,gua2020improving}
    or throughout training \cite{mahto2020multi}.

    \section{Time scale in RNNs under training}\label{sec:extention}

    The existing theory \cite{tallec2018can} interprets the rate of decay of the state
    as a time scale in leaky or gated RNNs (Section \ref{sec:preliminaries}) under the 
    assumption of zero input $\bx_t = 0$ for $t>t_0$.
    Furthermore, it assumes various extreme conditions on parameters such as
    $\bb=0$ and $\bU=0$ in (\ref{eq:LeakyRNN}).
    In order to justify the interpretation in practical situations,
    we need to deal with the following two issues;
    \begin{enumerate}
        \item inputs $\bx_t$ after time $t_0$, which have an effect on the state $\bh_t$,
              are successively given to a RNN, and
        \item the quantity $\tilde \bh_t$ representing new information is related to
              the previous state, which may have an effect
              on the remaining time of information the state has.
    \end{enumerate}
    In this section, we work around the first issue by reformulating the theory
    on time scale in RNNs.
    Our idea is to consider the Jacobian matrix of the state,
    rather than the state itself, to formulate the memory of RNN models.
    We further discuss the second issue by combining this reformulation with
    experimental observations.

    \subsection{Revisiting the time scale theory on RNNs}\label{subsec:revisit}

    First, we consider the following generalization of
    the theory on time scale in RNNs to admit successive inputs.
    The amount of information on $\bh_{t_0}$ that $\bh_t$ has can be quantified 
    as a variation of $\bh_t$ when $\bh_{t_0}$ is infinitesimally changed.
    Mathematically, this amounts to measuring the Jacobian matrix 
    $\frac{\partial \mathbf h_t}{\partial \mathbf h_{t_0}} (\mathbf h_{t_0})$
    with $\mathbf h_t$ considered as a function of $\bh_{t_0}$.
    To see how this new aspect naturally generalizes the existing theory,
    let us first discuss this idea with a leaky RNN (\ref{eq:LeakyRNN})
    (we treat a general case with gating mechanism later).
    We rewrite the solution of the ``free input" equation (\ref{eq:simple-decay})
    as $\frac{\partial \mathbf h_t}{\partial \mathbf h_{t_0}} (\mathbf h_{t_0}) = (1-\alpha)^{t-t_0}\mathbf I$.
    This equation indicates that the state $\mathbf h_{t_0}$ at time $t_0$ has
    exponentially decaying effects on the state $\mathbf h_t$ with characteristic time $\tau$.
    Note that this equation holds even when we admit inputs after $t_0$ and 
    a non-zero bias $\mathbf b$, that is, a solution $\mathbf h_t$ for a system
    \begin{align}\label{eq:decay-with-input}
        \mathbf h_t = (1-\alpha) \mathbf h_{t-1} + \alpha \tanh (\mathbf W \mathbf x_t + \mathbf b)
    \end{align}
    satisfies $\frac{\partial \mathbf h_t}{\partial \mathbf h_{t_0}} (\mathbf h_{t_0}) = (1-\alpha)^{t-t_0}\mathbf I$.
    This can be checked by the equation $\partial \mathbf h_t / \partial \mathbf h_{t-1} = (1-\alpha)\mathbf I$
    and the chain rule.
    Thus, formulating the memory of a state as a magnitude of the Jacobian matrix 
    $\partial \mathbf h_t / \partial \mathbf h_{t_0}$
    is also reasonable in an ``input existing" regime.
    Note that this is closely related to the vanishing gradient problem \cite{bengio1994learning,pascanu2013difficulty}.
    Since the gradient of loss function $L$ with respect to the state $\mathbf h_{t_0}$
    at time $t_0$ is calculated by $\frac{\partial L}{\partial \mathbf h_{t_0}} = 
    \frac{\partial \mathbf h_t}{\partial \mathbf h_{t_0}} \frac{\partial L}{\partial \mathbf h_t}$,
    the fast exponential decay in the Jacobian matrix $\partial \mathbf h_t / \partial \mathbf h_{t_0}$
    tends to lead to the extreme small gradient of loss function $\partial L / \partial \mathbf h_{t_0}$.

    The case for gated RNNs is similar.
    Ignoring recurrent weight matrices in (\ref{eq:gatedRNN}) by assuming
    $\partial \tilde \bh_t / \partial \bh_{t-1} = 0, \bU_f=0,$ and $\bU_i = 0$,
    we obtain the Jacobian matrix as
    $\partial \mathbf h_t / \partial \mathbf h_{t_0} = \prod_{s=t_0 +1}^{t} \mathbf f_t$.
    The decaying behavior of the Jacobian matrix is exactly the same as the state 
    under the ``free input'' regime in Section \ref{sec:preliminaries}.

    \begin{rem}
        It is not obvious what kind of value of a matrix $\partial \mathbf h_t / \partial \mathbf h_{t-1}$
        we should use to represent memory more clearly.
        Adopting the spectral norm or the spectral radius is a natural choice,
        which has indeed been taken previously to analyze the memory of general RNNs
        \cite{le2019Learning}.
        Considering memory as preserving the state {\it as it is}, 
        it might be more appropriate to deal with 
        how close to the identity matrix the Jacobian matrix $\partial \mathbf h_t / \partial \mathbf h_{t-1}$ is.
        We do not delve into this subtlety here, 
        as we are only interested in how models behave compared to the simplified ones
        like (\ref{eq:decay-with-input}) in realistic situations.
    \end{rem}

    \subsection{Observations on time scale in RNN under training}\label{subsec:observation}
    
    We move on to the second issue, that is,
    how to deal with the dependency of ignored terms such as $\tilde \bh_t$
    on the previous state $\bh_{t-1}$.
    This issue is more subtle than the first one,
    since when recurrent weight matrices are large,
    the Jacobian matrix $\partial \mathbf h_t / \partial \mathbf h_{t_0}$ gets unbounded
    and leaky units or forget gates might not represent the time scale in RNNs any more\footnote{This
    consideration is in a similar spirit to a study on chaotic 
    behavior in RNNs \cite{laurent2016recurrent},
    which observes trained gated RNNs from a dynamical system viewpoint.}.
    To tackle this, we first determine when the recurrent weight matrices are small enough to be ignored.

    To examine this closely, we again begin with a leaky RNN, whose Jacobian matrix of 
    1-step state update (\ref{eq:LeakyRNN}) is given by
    \begin{align}\label{eq:jacobi-leaky}
        \frac {\partial \bh_t}{\partial \bh_{t-1}} = (1-\alpha) \mathbf I - \alpha \bU^\top \mathbf D_t,
    \end{align}
    where $\mathbf D_t = \operatorname{diag} (\tanh'(\bU \bh_{t-1} + \bW \bx_t + \bb))
                       = \operatorname{diag} (\mathbf 1- \bh_t^2)$
    is a diagonal matrix given by a derivative of activation function $\tanh$.
    If $\bU$ is large enough relative to $\mathbf I$, e.g., $\bU = c\mathbf I$ 
    with $c\gg 1$, then we cannot ensure that
    the first term $(1-\alpha) \mathbf I$ controls the time scale of the state any more.

    The situation is more involved in the case of RNNs with a gating mechanism.
    For example, we consider a GRU, whose state transition is defined by 
    \begin{align}
        \mathbf h_t &= (\mathbf 1 - \mathbf z_t) \odot \mathbf h_{t-1} + \mathbf z_t \odot \tilde {\mathbf h}_t,  \\
        \mathbf z_t &= \sigma ( \mathbf U_z \mathbf h_{t-1} + \mathbf W_z \mathbf x_t + \mathbf b_z ),               \\
        \tilde {\mathbf h}_t &= \tanh ( \bU (\mathbf r_t \odot \bh_{t-1}) + \bW \bx_t + \bb ),                     \\
        \mathbf r_t &= \sigma ( \bU_r \bh_{t-1} + \bW_r \bx_t + \bb_r),
    \end{align}
    where $\bU_z,\bU_r \in \mathbb{R}^{n\times n},\bW_z,\bW_r \in \mathbb{R}^{n\times d}$, 
    and $\bb_z, \bb_r \in \mathbb{R}^n$ are additional parameters.
    The Jacobian matrix for this update is given as
    \begin{align}\label{eq:jacobi-gru}
        \frac{\partial\bh_t}{\partial\bh_{t-1}} 
        = &\operatorname{diag} (\mathbf 1 - \mathbf z_t)
          + \operatorname{diag} (\mathbf z_t) \frac{\partial \tilde {\mathbf h}_t}{\partial \bh_{t-1}}  \notag \\
          &+ \operatorname{diag}(\tilde \bh_t - \bh_{t-1}) \frac{\partial \mathbf z_t}{\partial \mathbf h_{t-1}}.
    \end{align}
    In addition to the ``time scale part" $\operatorname{diag} (\mathbf 1 - \mathbf z_t)$,
    there appear more terms involving the derivative of other functions such as 
    $\frac{\partial \tilde {\mathbf h}_t}{\partial \bh_{t-1}}$ and $\frac{\partial \mathbf z_t}{\partial \mathbf h_{t-1}}$.
    It is highly non-trivial to analyze when we can reasonably ignore each term.

    As we have seen, it is not straight-forward to take an analytical approach
    to determine when we can ignore the recurrent weight matrices.
    Luckily, we can utilize the generalized theory on time scale 
    in Section \ref{subsec:revisit} to deal with this issue.
    Namely, in order to ensure that terms other than the leading term (i.e., ``time scale part'') 
    such as $- \alpha \bU^\top \mathbf D_t$ in (\ref{eq:jacobi-leaky}) 
    and $\operatorname{diag} (\mathbf z_t) \frac{\partial \tilde {\mathbf h}_t}{\partial \bh_{t-1}} 
    + \operatorname{diag}(\tilde \bh_t - \bh_{t-1}) \frac{\partial \mathbf z_t}{\partial \mathbf h_{t-1}}$ 
    in (\ref{eq:jacobi-gru}) have a negligible effect on 
    the decay rate of the memory of the state,
    it is enough to directly check that the Jacobian matrix
    $\partial \bh_t / \partial \bh_{t_0}$ decays exponentially as time goes back.
    As the latter condition can be checked directly by the backpropagation method, 
    we can detect whether or not the recurrent weight matrices are ignorable
    without focusing on their values.
    This is a clear advantage over the conventional theory that interprets 
    the state itself as memory, while we use the Jacobian matrix instead.

    We now see how memory in RNNs behaves in practical situations.
    In general, the Jacobian matrix $\partial \mathbf h_t / \partial \mathbf h_{t_0}$ 
    is high dimensional and so requires much computational cost to compute.
    Thus, for simplicity, we treat the Euclidean norm of the gradient $||\partial L / \partial \mathbf x_t||$
    of cross-entropy loss with respect to an input at each time 
    in sequence classification tasks (Section \ref{sec:experiment}),
    instead of the Jacobian matrix.
    Since $\frac{\partial L}{\partial \mathbf x_t} =\frac{\partial \bh_t}{\partial \bx_t} \frac{\partial \bh_T}{\partial \bh_t} \frac{\partial L}{\partial \mathbf h_T}$
    holds where $T$ is the last time step to output prediction,
    norm of this gradient can be viewed as a proxy for the representation of memory
    in the state at prediction time $T$.
    If the norm of the gradient $||\partial L / \partial \mathbf x_t||$ 
    decreases exponentially during the backpropagation through time, 
    we can conclude that the gating mechanism
    or the leaky units indeed represent the time scale of the state.    
    We visualize how this value behaves while training for 
    leaky RNNs and LSTMs in Figure \ref{fig:grad_change}.
    We found that the gradient does not behave exponentially
    after learning.
    This indicates that there is a non-negligible effect of 
    recurrent weight matrices on the gradient,
    which might cause a gap between theoretical expectation
    and the actual behavior of the RNN model.
    In contrast, we observed that the input gradient decreases 
    exponentially with respect to time steps at initialization,
    regardless of tasks.
    This implies that at least at the initial learning phase,
    the leaky units and the forget gate function indeed
    represent the time scale in states.
    This phenomenon gives us a new insight on previously proposed
    initialization techniques for parameters in a forget gate 
    (Section \ref{sec:preliminaries}).
    Namely, the reason those initialization methods are effective is 
    probably that the time scale representation in forget gates
    is firmly valid at randomly initialized networks.
  
    \begin{figure}[t]
        \centering
        \includegraphics[width=0.4\paperwidth]{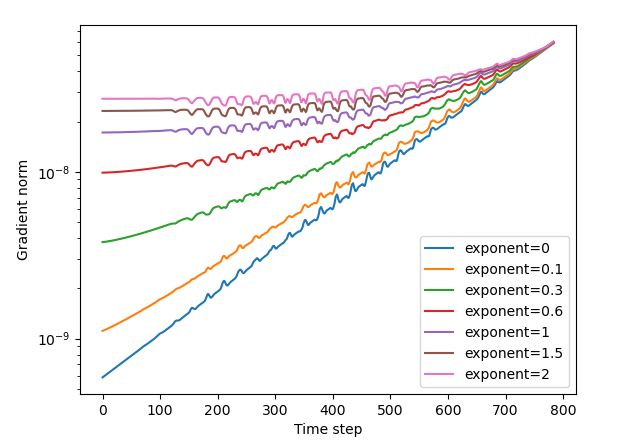}
        \caption{
            Euclidean norm of gradient of the cross-entropy loss with respect to
            an input at each time step $||\partial L / \partial \mathbf x_t||$
            for a randomly initialized models used in the sMNIST experiment 
            (Section \ref{sec:experiment}).
            The exponent is the value of the rate parameter $r$ in 
            our proposed method (Section \ref{sec:proposal}). 
            A leaky RNN model (exponent $r=0$) has an exponentially decaying gradient.
            With our proposed method applied, the gradient behaves polynomially.
            We initialized the weight matrix $\bU$ by a normal distribution 
            of mean 0 and standard deviation $0.1/\sqrt{n}$, where $n$ is the dimension of hidden states.    
        }
        \label{fig:grad_decay}
    \end{figure}

    \section{A method to represent longer time scale}\label{sec:proposal}

    Leaky units and forget gates model exponentially decaying memory
    by (\ref{eq:simple-decay})
    under settings where recurrent weight matrices are ignored.
    With such exponential decay,
    the memory is reduced by a constant factor at every constant time.
    For an RNN to hold memory for a much longer time,
    it is natural to expect memory to decay
    in a slower order, such as a polynomial order.
    In this section, we investigate how to incorporate 
    such slow decay into the model structure.
    This is done by introducing a higher degree term
    to the ordinary differential equation (ODE) counterpart
    of RNNs.

    We first derive the continuous version of RNN.
    We again consider a leaky RNN for simplicity.
    The difference of states is written as
    \begin{align}
        \mathbf h_t - \mathbf h_{t-1} = \alpha \bigl(\tanh(\mathbf U \mathbf h_{t-1} + \mathbf W \mathbf x_t + \mathbf b) - \mathbf h_{t-1}\bigr).
    \end{align}
    Considering a sufficiently small time step, we get an ODE
    \begin{align}\label{eq:ode-linear-decay}
        \frac{\mathrm{d}\mathbf h}{\mathrm{d} t} = \alpha \bigl(\tanh(\mathbf U \mathbf h (t) + \mathbf W \mathbf x (t) + \mathbf b) - \mathbf h (t)\bigr).
    \end{align}
    We are interested in the case when the recurrent weight matrix $\mathbf U$
    can be ignored, which we have confirmed for a model at initialization in 
    Section \ref{subsec:observation}.
    Setting the recurrent weight matrix $\mathbf U$ as $\bU = 0$ in (\ref{eq:ode-linear-decay}),
    we obtain a continuous counterpart of memory decay
    (discussed in Section \ref{subsec:revisit}) as
    \begin{align}
        \frac{\partial \bh(t)}{\partial \bh(t_0)} (\mathbf h(t_0)) = e^{-\alpha(t-t_0)} \mathbf I.
    \end{align}
    We aim to make this exponential decay polynomial,
    which is achieved by simply replacing a linear decay term $\bh(t)$
    in (\ref{eq:ode-linear-decay}) with a higher degree term.
    Formally, we consider an ODE for 1-dimensional state $h \in \mathbb R$ given by 
    \begin{align}
        \frac{\mathrm{d} h}{\mathrm{d} t} = - |h|^r h,
    \end{align}
    with the exponent $r \ge 0$ that determines the order of decay.
    Since this ODE is symmetric with respect to the origin $h=0$,
    we only consider the case where $h>0$.
    For $r=0$, this ODE has an exponential decaying solution,
    as discussed earlier.
    When $r>0$, the solution of this ODE is
    \begin{align}\label{eq:high-degree-solution}
        h (t) = \bigl( r(t-t_0) + \frac{1}{h(t_0)^r} \bigr)^{-1/r}.
    \end{align}
    Taking the derivative with respect to $h(t_0)$, we get
    \begin{align}\label{eq:high-degree-jacobian}
        \frac{\partial h(t)}{\partial h(t_0)} = 
        \bigl( 1 + h(t_0)^r (t-t_0) \bigr)^{-\frac{r+1}{r}},
    \end{align}
    which indicates a much slower decay of memory of the state $h$.
    Adopting this idea into (\ref{eq:ode-linear-decay}) leads to
    \begin{align}
        \frac{\mathrm{d} \bh}{\mathrm{d} t} = \alpha \bigl( \tanh(\mathbf U \mathbf h (t) + \mathbf W \mathbf x (t) + \mathbf b)
                             - |\bh(t)|^r\mathbf h (t) \bigr).
    \end{align}
    Going back to a discrete RNN model, we obtain a new architecture by
    \begin{align}\label{eq:proposed}
        \mathbf h_t &= \mathbf h_{t-1} + \alpha (\tilde{\mathbf h}_t - |\bh_{t-1}|^r \bh_{t-1}), \\
        \tilde{\mathbf h}_t &= \tanh(\mathbf U \mathbf h_{t-1} + \mathbf W \mathbf x_t + \mathbf b).
    \end{align}
    We treat the rate parameter $r>0$ as a hyperparameter.
    We propose this method to construct a new RNN models to represent 
    a much longer time scale.
    We have explained the method applied for a leaky RNN, but it naturally extends to
    general gated RNNs, replacing multiplication with a forget gate $\mathbf f_t \odot \bh_{t-1}$
    in (\ref{eq:gatedRNN}) by
    $\bh_{t-1} - (\mathbf 1 - \mathbf f_t) \odot |\bh_{t-1}|^r \bh_{t-1}$.
    This modification of the model structure causes to the increase in the computational complexity 
    by taking the absolute values and the power of the state.
    However, since the computational cost for such operations is relatively small compared to
    the matrix multiplication, the increase is negligible in the whole forward and backward computation.

    As we have seen in Section \ref{subsec:observation},
    the effectiveness of imposing the desired time scale of states
    might particularly have an effect in the initial learning phase.
    Therefore, we want to achieve the ``slow decaying memory" 
    in the proposed model at least at initialization.
    We visualize the input gradient on the proposed model with random initialization 
    by changing the rate parameter $r$ from 0 to 2 in Figure \ref{fig:grad_decay}.
    We see that the gradient decays polynomially instead of exponentially
    when $r>0$,
    which indicates that the proposed model has a longer memory than
    the baseline leaky RNN model ($r=0$).


    \begin{table*}[t]
        \caption{Statistics of Dataset and Training Setup}
        \label{tab:statistics}
        \centering
        \begin{tabular}{|c|rrrrr|rrrrr|}
            \hline
            Task & Train & Valid. & Test & Input dim. & Time steps &
            Hidden dim. & Optimizer & Learning rate & Gradient clip \cite{pascanu2013difficulty}
            & Batch size \\
            \hline
            sMNIST  & 50000 & 10000 & 10000 & 1 & 784 & 128 & RMSprop & \{1e-3, 5e-3, 1e-4\} & \{1, 5\} & 100 \\
            psMNIST & 50000 & 10000 & 10000 & 1 & 784 & 128 & RMSprop & \{1e-3, 5e-3, 1e-4\} & \{1, 5\} & 100 \\
            HAR     & 5881  & 1471  & 2947  & 9 & 128 & 64  & RMSprop & \{1e-3, 5e-3, 1e-4\} & \{1, 5\} & 100 \\
            \hline
        \end{tabular}
    \end{table*}
    
    \begin{table}[t]
        \caption{Test Accuracy}
        \label{tab:results}
        \centering
        \begin{tabular}{crrr}
            \hline
            & sMNIST & psMNIST & HAR \\
            \hline \hline
            Leaky RNN & 94.6 & 90.7 & 91.4 \\
            Ours    & 92.4 & 91.7 & 92.1 \\
            \hline
        \end{tabular}
    \end{table}

    \section{Experimental evaluation}\label{sec:experiment}

    We conduct experiments to determine the effectiveness of our proposed method.
    We evaluate a baseline leaky RNN model (\ref{eq:LeakyRNN}) (Leaky RNN) and 
    a model modified with our proposed method (\ref{eq:proposed}) (Ours) on two sequence classification tasks.
    One is a pixel-by-pixel image recognition task,
    which is often used to benchmark how RNN models capture
    long-term dependencies.
    The other is a human action recognition task, which
    is taken as a more practical task for RNN applications.

    \subsection{Experiment setting}

    The statistics and hyperparameters used for each task are shown in Table \ref{tab:statistics}.
    After each epoch of training, we record the validation loss on the validation data.
    We train each model for 200 epochs on each task, reducing the learning rate
    by half after 100 and 150 epochs.
    We perform random initialization and training four times on each setting.
    Then, we evaluate the accuracy for the test data on the model that has
    the lowest validation loss among all hyperparameters, epochs, and initialization. 
    
    The performance of both Leaky RNN and Ours strongly depends on the initialization of 
    the time scale parameter $\alpha$.
    It is recommended to set $\alpha$ proportionally to $1/T$ for the sequance length $T$
    to easily adjust the time scale of the state to that of the data \cite{tallec2018can,kusupati2019fastgrnn}.
    Thus, we test the initialization by $\alpha = 1/T, 5/T$, and $25/T$
    and choose the best-performing one according to the validation loss.

    In the proposed method (\ref{eq:proposed}),
    we use the following setting.
    For the rate parameter $r$, we use $r=2$,
    which simplifies the implementation as $|\bh_{t-1}|^r \bh_{t-1} = \bh_{t-1}^3$.
    We initialize the recurrent weight matrix $\bU$ by a normal distribution 
    of mean 0 and standard deviation $0.1/\sqrt{n}$, where $n$ is the dimension of hidden states
    (see Figure \ref{fig:grad_decay}).

    Our computational setup is the following:
    CPU is Intel Xeon Silver 4214R 2.40GHz, the memory size is 512 GB, 
    and GPU is NVIDIA Tesla V100S.

    \begin{figure}[t]
        \centering
        \includegraphics[width=0.4\paperwidth]{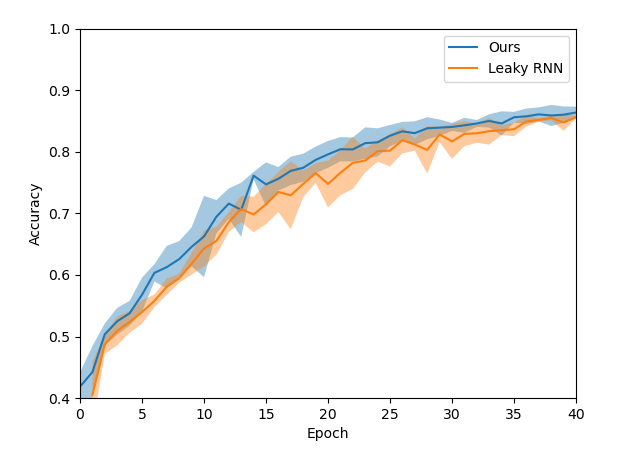}
        \caption{
            Mean validation accuracy over 4 independent learning iterations of 
            models under the best performing training setup 
            on psMNIST task (Section \ref{sec:experiment}).
            The shaded area shows the standard deviation.
        }
        \label{fig:psMNIST_acc}
    \end{figure}

    \begin{figure}[t]
        \centering
        \includegraphics[width=0.4\paperwidth]{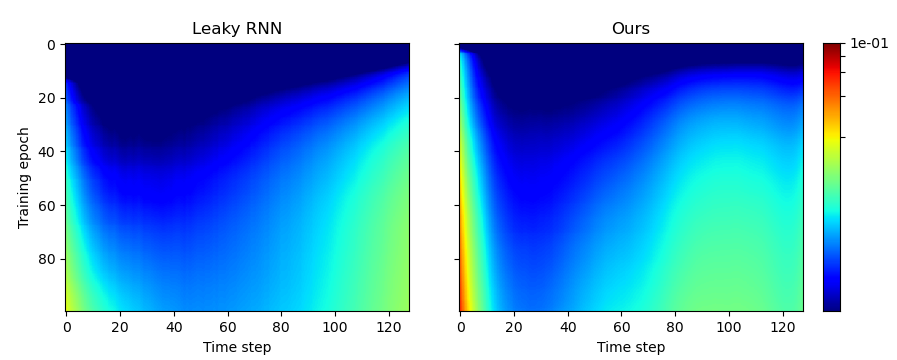}
        \caption{
            Euclidean norm of gradient of cross-entropy loss
            with respect to inputs on HAR task.
            Color bar is set as log scale.
            Ours (right) has a wider range of time steps 
            with large gradient than Leaky RNN (Left)
            in the early stage of learning.
        }
        \label{fig:grad_HAR}
    \end{figure}

    \subsection{Sequential MNIST (sMNIST)}

    We evaluate models on two pixel-by-pixel image recognition tasks: 
    sequential MNIST\footnote{http://yann.lecun.com/exdb/mnist/
    } (sMNIST), and permuted sequential MNIST (psMNIST)
    \cite{le2015simple}.
    In these tasks, an image of size $28\times 28$ is treated as a sequance of 
    1-dimensional pixel values of length $28 \times 28 = 784$.
    An RNN is given a pixel value as an input at each time step.
    After that, it predicts the label of the image.
    Since the RNN needs to utilize information on distant pixels to classify images correctly,
    these tasks have been used to test ability of RNNs to learn how to capture
    long-term dependencies of data.
    In the sMNIST task, the pixels are input to an RNN in an ordered way,
    from left-to-right and top-to-bottom.
    To introduce more complex tempral dependencies, a fixed random permutation
    is applied to pixels in the psMNIST task.

    We show the results in Table \ref{tab:results} (sMNIST/psMNIST).
    While the baseline Leaky RNN shows a higher test accuracy on the sMNIST task,
    Ours outperforms the baseline on the psMNIST task.
    While RNN models can classify most images correctly
    by using relatively short-term dependencies of data on the sMNIST task,
    they need to exploit information on far distant inputs on the psMNIST task.
    Therefore, we conclude that the proposed method
    can improve the accuracy especially
    on the data with more complex and longer time scales.
    
    Our proposed method builds on the hypothesis that 
    imposing a proper time scale on an RNN at the initial phase of training
    improves the learnability of the model.
    This suggests that the proposed method may have 
    a particular effect on the accuracy at the initial phase of training.
    To examine this, we visualize the transition of the validation loss while training
    in Figure \ref{fig:psMNIST_acc}.
    We observe that our method indeed improves the performance
    especially at the earlier stage of training.

    \subsection{Human action recognition (HAR)}

    We next evaluate the models on more practical datasets,
    that contains three types of sensor data
    for the x, y, and z axes labeled with
    six types of human action\footnote{https://archive.ics.uci.edu/ml/datasets/human+activity+recognition+using+
    smartphones} \cite{anguita2012human}:
    walking, walking upstairs, walking downstairs,
    sitting, standing, and laying.
    We preprocessed the data and the labels in the same way as
    the previous work \cite{kusupati2019fastgrnn} to make this task
    into a binary classification on normalized data.

    We show the results on Table \ref{tab:results} (HAR).
    Ours gives better test accuracy than the baseline.
    This result demonstrates that our proposed method improves 
    the learnability of RNNs for high-dimensional data with complex time scales.
    We further visualize the gradient of cross-entropy loss with respect to inputs 
    for each model in Figure \ref{fig:grad_HAR}.
    We observe that Ours takes large values on a wider range of time steps
    especially at the early stage of training ($\sim 60$ epochs).
    This indicates that modeling slow decaying memory helps the RNNs
    to capture complex temporal dependencies over the whole sequence,
    which results in an improvement of accuracy.

    \begin{figure}[t]
        \centering
        \includegraphics[width=0.4\paperwidth]{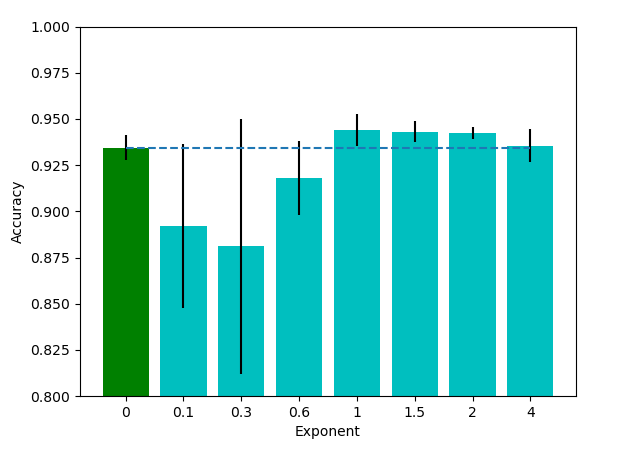}
        \caption{
            Validation accuracy on HAR task for various rate parameters $r$ in
            (\ref{eq:proposed}).
            Mean and standard error are taken over four independent training results.
            The exponent $r=0$ case (green) corresponds to the baseline Leaky RNN.
        }
        \label{fig:acc_exponents}
    \end{figure}

    \subsection{Sensitivity to the rate parameter}

    Our proposed method contains the rate parameter $r$ as a hyperparameter.
    Taking a limit $r \to 0$, our method reduces to conventional models.
    In contrast, a larger $r$ corresponds to modeling the memory decaying more slowly.
    In this subsection, we investigate how change in $r$ affects the learnability 
    of RNNs.
    We test the proposed method applied to a leaky RNN, changing $r$ from 0 to a
    larger value, under the same setting as the previous evaluation on the HAR task.
    When $r$ is smaller than 1, 
    we found that the training is unstable, and it tends to diverge 
    in few epochs even with gradient clipping 
    \cite{pascanu2013difficulty} (Figure \ref{fig:acc_exponents}).
    This might be caused by numerical instability in forward (\ref{eq:high-degree-solution})
    or backward (\ref{eq:high-degree-jacobian})
    computation for the polynomially decaying quantity, or in its interaction with 
    the recurrent weight matrix $\bU$.
    Interestingly, the results show that the learning dynamics in the proposed method
    behaves discontinuously with respect to the limit $r\to 0$,
    despite the continuity of the gradient behavior at initialization
    (Figure \ref{fig:grad_decay}).
    On the other hand, for a large value of $r$, e.g., $r=4$, we found that 
    the training converges more slowly and results in a lower accuracy 
    than the $1\le r \le 2$ cases (Figure \ref{fig:acc_exponents}).
    This result indicates that retaining too much memory might
    prevent smooth learning.
    This is consistent with previous findings 
    that it is desirable to model ``forgetting''
    appropriately for improving the learnability of RNNs
    \cite{gers1999learning,jing2019gated}.
    Therefore, it is important to choose a proper rate parameter $r$
    to achieve higher accuracy.
    We hypothesize that taking around $r = 2$ might generally work,
    as the higher accuracy is also obtained in the psMNIST task 
    with $r = 2$ compared to the baseline.
    More detailed analysis on the instability and further investigation
    on applications to general gated RNNs will be a future work.

    \section{Conclusion}\label{sec:conclusion}

    In this paper, we extended the existing theory on 
    temporal representation of the forget gate function in
    gated RNNs to make it applicable in practical situations.
    We empirically showed that gated RNNs typically behave
    as the theory predicts at least at the initial phase of learning,
    which is in good agreement with the previously proposed initialization methods.
    We proposed a method to change the RNN structure to 
    improve learnability for data with long-term dependencies.
    Finally, we demonstrated the effectiveness of our method
    on real-world datasets.
    The results highlight the importance of theoretical
    modeling and of understanding the behavior of RNNs in 
    practical settings.

    \IEEEtriggeratref{17}
    \bibliographystyle{IEEEtran}
    \bibliography{IEEEabrv,ref}

\end{document}